\documentclass{article}

\usepackage[preprint]{neurips_2025}
\workshoptitle{Published in the proceedings of the 39th Conference on Neural Information Processing Systems (NeurIPS 2025) Workshop: Scaling Environments for Agents (SEA). Additionally accepted for presentation in NeurIPS 2025 Workshop: Embodied World Models for Decision Making}

\usepackage[utf8]{inputenc} % allow utf-8 input
\usepackage[T1]{fontenc}    % use 8-bit T1 fonts
\usepackage[colorlinks=true, allcolors=blue]{hyperref}
\usepackage{url}            % simple URL typesetting
\usepackage{booktabs}       % professional-quality tables
\usepackage{amsfonts}       % blackboard math symbols
\usepackage{nicefrac}       % compact symbols for 1/2, etc.
\usepackage{microtype}      % microtypography
\usepackage{xcolor}         % colors

\usepackage{graphicx}
\usepackage{amsmath}
\usepackage{listings}
\usepackage{enumitem}
\usepackage{algorithm}
\usepackage[noend]{algpseudocode}

\definecolor{codegray}{gray}{0.5}
\definecolor{codeblue}{rgb}{0.25,0.5,0.75}

\title{The Physical Basis of Prediction: World Model Formation in Neural Organoids via an LLM-Generated Curriculum}

\author{
Brennen Hill\\
Department of Computer Science\\
University of Wisconsin-Madison\\
Madison, WI 53706 \\
\texttt{bahill4@wisc.edu} \\
}

\begin{document}

\maketitle

\begin{abstract}
The capacity of an embodied agent to understand, predict, and interact with its environment is fundamentally contingent on an internal world model. This paper introduces a novel framework for investigating the formation and adaptation of such world models within a biological substrate: human neural organoids. We present a curriculum of three scalable, closed-loop virtual environments designed to train these biological agents and probe the underlying synaptic mechanisms of learning, such as long-term potentiation (LTP) and long-term depression (LTD). We detail the design of three distinct task environments that demand progressively more sophisticated world models for successful decision-making: (1) a conditional avoidance task for learning static state-action contingencies, (2) a one-dimensional predator-prey scenario for goal-directed interaction, and (3) a replication of the classic Pong game for modeling dynamic, continuous-time systems. For each environment, we formalize the state and action spaces, the sensory encoding and motor decoding mechanisms, and the feedback protocols based on predictable (reward) and unpredictable (punishment) stimulation, which serve to drive model refinement. In a significant methodological advance, we propose a meta-learning approach where a Large Language Model (LLM) automates the generative design and optimization of experimental protocols, thereby scaling the process of environment and curriculum design. Finally, we outline a multi-modal evaluation strategy that moves beyond task performance to directly measure the physical correlates of the learned world model by quantifying synaptic plasticity at electrophysiological, cellular, and molecular levels. This work bridges the gap between model-based reinforcement learning and computational neuroscience, offering a unique platform for studying embodiment, decision-making, and the physical basis of intelligence.
\end{abstract}

\section{Introduction}
The development of sophisticated world models has emerged as a cornerstone of modern embodied artificial intelligence, enabling agents to reason, plan, and act effectively in complex environments \citep{sutton2018reinforcement}. While progress in AI has been propelled by the concurrent scaling of model architectures, datasets, and computational resources \citep{brown2020language}, the crucial role of the environment itself as a driver of intelligence remains a frontier for exploration. The structure, complexity, and interactivity of an agent's world directly constrain and shape the internal models it can acquire. Rich, dynamic environments provide the necessary data streams for agents to move beyond passive prediction and static imitation, fostering the development of robust, goal-directed behaviors through continuous interaction.

This paper explores these principles through the lens of a novel and powerful agent substrate: living neural organoids. Recent breakthroughs in stem cell biology \citep{lancaster2013cerebral} and neuro-interfacing technologies have converged to create the field of Organoid Intelligence (OI), where three-dimensional human neural cultures are embodied within simulated, interactive worlds \citep{smirnova2023organoid, kagan2022vitro}. These biological agents, derived from human cells, present an unprecedented opportunity to study the formation of learned world models not as an abstract computational process, but as a tangible biological phenomenon rooted in the physical principles of synaptic plasticity.

Here, we present a comprehensive framework for designing, scaling, and evaluating virtual environments specifically tailored to induce and investigate learning in organoid-based agents. We make three primary contributions. First, we introduce a curriculum of three distinct environments, each designed to probe a different facet of decision-making and world modeling, ranging from simple state avoidance to dynamic, long-horizon interception tasks. We formalize the complete closed-loop system, including sensory state encoding, motor action decoding, and a reinforcement mechanism grounded in neuroscientific principles. Second, we propose a novel approach for evaluating an agent's learned world model, moving beyond behavioral metrics to directly measure the underlying biophysical changes in synaptic strength, namely Long-Term Potentiation (LTP) and Long-Term Depression (LTD). This allows us to correlate task proficiency with the physical reorganization of the agent's neural architecture. Third, we introduce a generative, meta-learning framework that leverages a Large Language Model (LLM) to automate the design and optimization of experimental protocols. This positions the LLM as a tool for goal-directed environment generation, enabling a new paradigm for scaling and accelerating research into embodied intelligence. By wedding model-based reinforcement learning concepts with a physical biological system, our work offers a unique platform to advance our understanding of how embodied agents learn to understand, predict, and interact with their world.

\section{Related work}
The concept of embodying living neural tissue within artificial environments to study learning and adaptation has a rich history. Foundational work with dissociated 2D neuronal cultures on multi-electrode arrays (MEAs) established the principle of \textit{in vitro} embodiment, demonstrating that neurally-controlled animats could learn to navigate simulated worlds and exhibit goal-directed behavior when provided with a closed-loop system of sensory feedback and motor control \citep{demarse2001neurally}. This research provided the first proof-of-concept that biological neural networks, de-coupled from a biological body, could form instrumental relationships with an external, virtual world.

The evolution from flat, 2D cultures to 3D neural organoids marks a profound leap in biological fidelity \citep{lancaster2013cerebral}. Organoids leverage the principles of developmental self-organization \citep{Zhang2022} to recapitulate the complex cytoarchitecture, cellular diversity, and developmental trajectories of the human brain. This methodology has produced both unguided cerebral organoids, which stochastically generate multiple brain regions to model inter-regional interactions \citep{Lancaster2014Generation}, and guided organoids, which use morphogens to direct differentiation toward specific brain regions (e.g., cortex) for improved reproducibility \citep{Shaker2022, Pavlinek2025}. This offers a far richer substrate for investigating the network-level phenomena that underpin complex decision-making and the formation of internal models. Recent work has successfully demonstrated that these organoid-based systems can learn to perform tasks, such as playing a simplified version of the game Pong, by harnessing their inherent plasticity \citep{kagan2022vitro}.

Our framework builds directly upon these advancements by grounding the learning mechanism in the free-energy principle \citep{friston2010free}. This theoretical framework posits that self-organizing systems, including the brain, act to minimize prediction error or surprise. Within this paradigm, predictable sensory stimuli can be interpreted as an intrinsic reward, reinforcing the policies that led to them, while unpredictable, high-entropy stimuli act as an aversive signal or punishment, indicating a failure of the agent's internal world model. This provides a powerful, neuro-plausible mechanism for implementing model-based reinforcement learning, where feedback directly drives the refinement of the agent's predictive model of its environment.

Furthermore, our work introduces a significant methodological innovation by employing an LLM to automate the design of experimental protocols. This aligns with an emergent trend of leveraging AI for autonomous scientific discovery. Such systems have already demonstrated success in optimizing complex processes in domains like chemistry and materials science \citep{boiko2023autonomous}. By applying this paradigm to the training of embodied agents, we propose a system where the LLM acts as a meta-controller, designing the very worlds and curricula in which the biological agent learns. This approach addresses leveraging the knowledge within LLMs to guide agent decision-making, extending the concept to guide the entire learning and discovery process itself.

\section{Methods}

\subsection{Multi-electrode array interface}
All proposed experiments are built upon a multi-electrode array (MEA) platform, which provides the critical bidirectional interface for embodying the biological agent. This platform must be capable of concurrent, high-fidelity stimulation and recording from the neural culture. For simplicity and reproducibility, we assume a setup with at least four distinct electrode groups (A, B, C, D), which can be spatially organized across the array. Each group can consist of one or more electrodes. Groups A and B are designated for recording spontaneous and evoked neural activity, serving as the agent's motor output channels. Groups C and D are dedicated to delivering patterned electrical stimuli, functioning as the agent's sensory input channels. However, the scaling of environments and the fine-grained evaluation of the world model ideally require more advanced interfaces. These include high-density CMOS-based MEAs (HD-MEAs), which offer thousands of electrodes for high-resolution field potential imaging \citep{Remi2025}, or flexible mesh and 3D-pillar MEAs that achieve superior volumetric integration and signal quality with 3D tissues \citep{MCDONALD2023115223}. This architecture forms the physical link between the virtual environment and the biological substrate.

\subsection{Biological substrates and embodiment}
The primary biological substrate for this work is the human neural organoid, chosen for its structural and cellular complexity \citep{lancaster2013cerebral}. The success of any training protocol is critically contingent on the developmental maturity and long-term viability of these organoids.

\textbf{Maturation and Culture Requirements:} Neural organoids require a protracted \textit{in vitro} culture period, typically a minimum of 60-90 days, to develop the cellular diversity (including neurons, astrocytes, and other glia \citep{Porciuncula2021}) and functional network activity, such as synchronized bursting, required for learning \citep{fair2020electrophysiological, Giandomenico2021Generation}. During this extended maturation, organoids must be maintained in a dynamic culture system, such as on an orbital shaker or in a spinning bioreactor. This is essential to enhance nutrient and oxygen diffusion, preventing the formation of a hypoxic or necrotic core that compromises tissue health and experimental reproducibility \citep{Zhang2022}. Long-term viability is further supported by specialized, serum-free media formulations (e.g., BrainPhys-based media) supplemented with neurotrophic factors like BDNF and GDNF \citep{Zhang2022, fair2020electrophysiological}.

\textbf{Protocol Choice and Reproducibility:} The experiments proposed can utilize organoids from either \textit{guided} or \textit{unguided} differentiation protocols. Guided protocols, which use exogenous morphogens to direct differentiation towards a specific regional identity like the dorsal forebrain \citep{Shaker2022}, are preferred for tasks requiring high reproducibility and lower inter-organoid variability. Unguided protocols, which rely on intrinsic self-organization, generate a more complex and heterogeneous mix of brain regions \citep{Lancaster2014Generation}, but this stochasticity presents a significant challenge for controlled learning experiments. Therefore, a semi-guided protocol \citep{Fitzgerald2024} or the use of guided cortical organoids represents the most promising balance of complexity and consistency.

\textbf{Integration with the MEA:} For all experiments, healthy, mature organoids must be selected and carefully plated onto the MEA surface, which is pre-treated with adhesion-promoting molecules like Poly-D-Lysine and Laminin \citep{Hales2010}. A stable, high-quality electrical interface is achieved by using surface tension to gently attach the organoid \citep{Remi2025} and allowing the system to acclimatize at 37°C and 5\% CO$_2$ for at least 45-90 minutes before recording to establish a stable baseline \citep{fair2020electrophysiological}. While this paper focuses on whole organoids, an alternative for higher-throughput screening of the LLM-generated protocols would be to use 2.5D dissociated organoid cultures, which sacrifice 3D architecture for improved homogeneity \citep{Pavlinek2025}.

\subsection{Inducing world model formation via predictive coding and reinforcement}
The core learning mechanism of our framework is predicated on the principles of model-based reinforcement learning (RL), operationalized through the neuro-plausible lens of the free-energy principle \citep{friston2010free}. This principle suggests that biological agents inherently act to minimize the discrepancy between their predictions about the world and the sensory evidence they receive, a quantity known as surprise or prediction error. In our framework, we leverage this concept by designing feedback signals to be either highly predictable (low surprise) or highly unpredictable (high surprise), serving as functional proxies for positive and negative reinforcement, respectively. This mechanism provides the essential error signals required to drive synaptic plasticity, the biological substrate for updating the agent's internal world model.

\textbf{Reward (Model Confirmation).} A reward signal, delivered upon achieving a desirable state (e.g., capturing prey), is implemented as a predictable, low-entropy electrical stimulus. A concrete implementation involves delivering a consistent, low-frequency sinusoidal wave across all stimulation electrodes. This spatially and temporally predictable sensory input minimizes surprise, serving as a confirmation signal that reinforces the agent's preceding policy and strengthens its internal model of successful state-action sequences. An alternative biological implementation involves the targeted delivery of dopamine via UV light-controlled uncaging. This is achieved by perfusing the culture with a biologically inactive 'caged' dopamine compound; a focused pulse of light then cleaves the caging group, releasing active dopamine at a precise spatiotemporal location \citep{Davies2019}. This directly activates the brain's reward pathways, which are known to encode reward prediction errors \citep{Wolfram1998} and are crucial for inducing long-term potentiation \citep{schultz1997neural}.

\textbf{Punishment (Model Violation).} Conversely, a punishment signal, corresponding to an undesirable outcome (e.g., entering an aversive zone), is delivered as an unpredictable, high-entropy stimulus. This is implemented by applying a white-noise electrical signal, characterized by random amplitude and frequency, to a randomly selected subset of stimulation electrodes. This chaotic and surprising sensory input functions as a powerful error signal, indicating a violation of the agent's predictive model. The objective is to drive the network to update its policy to avoid states and actions that lead to such high-surprise, model-violating outcomes, providing a direct test of computational theories like the Free Energy Principle \citep{e24020301}.

This bipolar feedback scheme directly translates the abstract principles of reinforcement learning into tangible biophysical drivers of change. By selectively confirming successful policies with predictable rewards and signaling model failures with unpredictable punishments, the framework compels the biological neural network to iteratively build and refine an internal model that can better predict and control its virtual environment.

\textbf{Encoding State}. To enable the formation of a world model, the agent must receive information about the state of its environment. The agent's location, or the location of objects in the virtual world, is encoded through patterned electrical stimulation. For a discrete state space with 8 locations, each location can be mapped to a specific stimulation electrode within a sensory group. For continuous or higher-dimensional states, information can be rate-coded, where the frequency of stimulation on a given electrode corresponds to a variable like distance or position along an axis. This provides the raw sensory data from which the agent must construct its model.

\subsection{Evaluating the embodied world model: From task performance to synaptic correlates}
A profound advantage of employing a biological agent is the ability to move beyond simplistic behavioral metrics and directly measure learning as the physical reorganization of the agent's underlying neural substrate. Our framework proposes a multi-scale evaluation strategy to quantify this plasticity, allowing us to correlate improvements in decision-making with the specific biological changes that constitute the learned world model.

At the functional network level, we can probe synaptic efficacy by measuring field excitatory postsynaptic potentials (fEPSPs) before and after training epochs. By delivering a controlled test pulse to a stimulation electrode and recording the evoked potential in a downstream population, the slope of the fEPSP serves as a robust proxy for synaptic strength. A sustained increase in this slope post-training is the canonical electrophysiological signature of long-term potentiation (LTP), while a decrease signifies long-term depression (LTD). This plasticity can be actively probed by applying specific induction protocols, such as high-frequency tetanic stimulation (e.g., 100 Hz) to induce LTP or prolonged low-frequency stimulation (e.g., 1 Hz) to induce LTD \citep{Nabavi2014}. This provides a direct, quantifiable link between the agent's interactive experience and the Hebbian plasticity mechanisms thought to underlie learning and memory \citep{bliss1973long, malenka2004ltp}.

To resolve how neural populations encode task variables and represent the world model at a finer scale, we can employ optical imaging. In organoids engineered to express Genetically Encoded Calcium Indicators (GECIs) like GCaMP \citep{chen2013ultrasensitive}, often delivered via AAV transduction \citep{Giandomenico2021Generation}, two-photon microscopy enables longitudinal tracking of the activity of hundreds of individual neurons. This approach is ideally performed simultaneously with MEA recording to correlate cellular-level calcium transients with network-level electrical events \citep{Irwin2013}. This technique allows us to observe the evolution of network dynamics as learning progresses, potentially identifying the emergence of stable neural ensembles that represent specific states, actions, or environmental features. Changes in the synchrony, firing rate, and spatial organization of these calcium transients can elucidate how the agent's internal model is formed, stored, and refined through goal-directed interaction.

Finally, at the conclusion of an experiment, post-hoc molecular analyses can reveal the structural correlates of the observed functional changes. Immunohistochemistry can be used to visualize the molecular machinery of plasticity. For example, staining for the differential trafficking of AMPA and NMDA receptor subunits can expose changes in synaptic receptor density, a key mechanism underlying the expression of LTP and LTD \citep{malenka2004ltp}. This can be complemented by staining for presynaptic (e.g., Synapsin-1) and postsynaptic (e.g., PSD-95) markers to quantify changes in synapse density \citep{Yakoub2019}, general cell-type markers (e.g., MAP2 for neurons, GFAP for astrocytes) to assess culture health, and activity-dependent immediate early genes like c-Fos to identify the specific neuronal ensembles that were recently active \citep{heo2023experience}. By integrating these electrophysiological, optical, and molecular benchmarks, we can construct a comprehensive portrait of learning that bridges behavioral performance with the fundamental principles of neural adaptation, offering a ground-truth look into the learned world model.

\section{Environments}
We propose a curriculum of three virtual environments, each designed to incrementally increase the demands on the agent's world modeling and decision-making capabilities.

\subsection{Environment 1: Conditional avoidance}
\begin{figure}[htbp]
\centering
\includegraphics[width=0.9\linewidth]{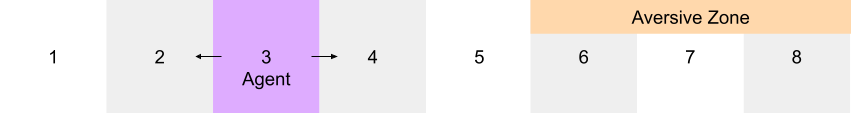} % Placeholder figure file
\caption{Diagram of the Conditional Avoidance environment. The agent, a neural organoid, is embodied in a 1D world. Its decision to move left or right is decoded from neural activity. Entering the aversive zone triggers unpredictable punishment, driving the agent to learn a world model where that region is associated with negative outcomes.}
\label{fig:figure1}
\end{figure}

This foundational experiment implements a classic aversive conditioning task, designed to test the agent's ability to build a simple world model that associates a specific region of its state space with negative outcomes and to develop a policy to actively avoid it. This environment, inspired by foundational experiments in learning theory \citep{Miller1948fear}, probes the most basic form of goal-directed behavior: avoidance of harm. A schematic is shown in Figure \ref{fig:figure1}.

\subsubsection{Protocol description}
\begin{enumerate}[label=\arabic*., wide, labelindent=0pt]
\item \textbf{State Representation and Sensory Encoding:} The agent is embodied as a point in a one-dimensional grid of 8 positions. Its current location is continuously conveyed via a unique spatio-temporal electrical stimulation pattern delivered across sensory electrode groups C and D. This provides the organoid with proprioceptive information about its virtual location. The state space is partitioned into a safe zone (positions 1-5) and an aversive zone (positions 6-8).

\item \textbf{Action Decoding and Execution:} The agent's intended movement is decoded by comparing the aggregate spike counts recorded from motor output groups A and B over a discrete time window $T$. If the integrated activity from group A exceeds that of group B, the agent's virtual position is decremented (action: move left). Conversely, if group B's activity is dominant, the position is incremented (action: move right). This establishes a simple, differential mapping from network activity to a binary action space.

\item \textbf{Feedback and State Transition:} The environment delivers feedback contingent on the agent's state. Upon entering the aversive zone, the agent receives an unpredictable, high-entropy electrical stimulus (punishment). The intensity of this stimulus is scaled with the depth of incursion (e.g., amplitude at position $8 > 7 > 6$), providing a gradient that can accelerate learning. Conversely, successfully remaining in the safe zone for a sustained duration of $Z$ timesteps triggers a predictable, low-entropy stimulus (reward).
\end{enumerate}

\begin{lstlisting}[caption=Pseudocode for conditional avoidance task, captionpos=b, label=lst:avoidance]
Initialize environment and agent interface
initialize_platform(groups_A, B, C, D)
agent_position = random.choice(columns_1_to_5)
safe_zone_timer = 0

Main experimental loop
while experiment_running:
# 1. State representation (sensory input)
encode_position_as_stimulus(groups_C_D, agent_position)

# 2. Action decoding (motor output)
spikes_A = record_spikes(group_A, window=T_ms)
spikes_B = record_spikes(group_B, window=T_ms)

if spikes_A > spikes_B:
    agent_position = max(1, agent_position - 1) # Move left
elif spikes_B > spikes_A:
    agent_position = min(8, agent_position + 1) # Move right

# 3. Feedback and state transition
if agent_position in columns_1_to_5: # In safe zone
    safe_zone_timer += 1
    if safe_zone_timer >= Z:
        deliver_reward(groups_C_D) # Predictable stimulus
        safe_zone_timer = 0
else: # In aversive zone
    safe_zone_timer = 0
    punishment_amplitude = calculate_amplitude(agent_position)
    deliver_punishment(groups_C_D, amplitude=punishment_amplitude)
\end{lstlisting}

\subsubsection{Environmental scaling and long-horizon planning}
\begin{itemize}
\item \textbf{Dimensionality:} The task's complexity can be significantly increased by scaling the world from a 1D line to a 2D grid or a 3D volume. A 2D environment requires a more sophisticated world model representing an (x, y) state, which could be encoded using separate electrode groups or frequency channels for each axis. The action space must expand to four cardinal directions, demanding a more complex motor decoding scheme, such as using four distinct recording sites or interpreting more complex spatio-temporal firing patterns. A pseudocode implementation is provided in Listing \ref{lst:avoidance_2d}.

\begin{lstlisting}[caption=Pseudocode for 2D conditional avoidance, captionpos=b, label=lst:avoidance_2d]
Initialize a 10x10 grid environment
GRID_SIZE = 10
agent_pos = (random.randint(0, GRID_SIZE-1), random.randint(0, 4)) # Start in safe half
aversive_zone_x = range(GRID_SIZE // 2, GRID_SIZE) # Right half is aversive

Assume 4 recording sites (A_up, A_down, B_left, B_right)
initialize_platform(groups_A_up, A_down, B_left, B_right, C, D)

while experiment_running:
# 1. State representation (2D)
encode_position_stimulus(group_C, axis='x', value=agent_pos[0])
encode_position_stimulus(group_D, axis='y', value=agent_pos[1])

# 2. Action decoding (4-way movement)
spikes_up = record_spikes(group_A_up, window=T_ms)
spikes_down = record_spikes(group_A_down, window=T_ms)
spikes_left = record_spikes(group_B_left, window=T_ms)
spikes_right = record_spikes(group_B_right, window=T_ms)

action = determine_dominant_action(spikes_up, spikes_down, spikes_left, spikes_right)

# Update position based on decoded action
x, y = agent_pos
if action == 'UP':    y = min(GRID_SIZE - 1, y + 1)
if action == 'DOWN':  y = max(0, y - 1)
if action == 'LEFT':  x = max(0, x - 1)
if action == 'RIGHT': x = min(GRID_SIZE - 1, x + 1)
agent_pos = (x, y)

# 3. Feedback based on 2D zone
if agent_pos[0] in aversive_zone_x:
    punishment_amplitude = calculate_amplitude(agent_pos[0])
    deliver_punishment(groups_C_D, amplitude=punishment_amplitude)
else:
    deliver_reward(groups_C_D) # Simplified: reward for being in safe zone
\end{lstlisting}

\item \textbf{Zone Complexity and Planning:} The simple binary safe/aversive partition can be replaced with more complex spatial arrangements that require rudimentary planning. For example, the agent could be rewarded for navigating a narrow, winding path or for solving a simple maze defined by aversive boundaries. This challenges the agent to build a world model that represents an extended spatial policy rather than a single boundary. See Listing \ref{lst:avoidance_maze}.

\begin{lstlisting}[caption=Pseudocode for maze navigation task, captionpos=b, label=lst:avoidance_maze]
Define a maze as a grid where 1=path, 0=wall
maze_layout = [
[1, 1, 1, 0, 1],
[0, 1, 0, 0, 1],
[1, 1, 1, 1, 1],
[1, 0, 1, 0, 1],
[1, 1, 1, 0, 1]
]
start_pos, goal_pos = (0, 0), (4, 4)
agent_pos = start_pos

while experiment_running:
# State and Action decoding remain similar to 2D avoidance
encode_position_stimulus(group_C, axis='x', value=agent_pos[0])
encode_position_stimulus(group_D, axis='y', value=agent_pos[1])

action = decode_4way_movement(groups_A_B)
next_pos = calculate_next_pos(agent_pos, action)

# 3. Feedback and state transition based on maze structure
x, y = next_pos
if maze_layout[y][x] == 1: # Valid path
    agent_pos = next_pos
    if agent_pos == goal_pos:
        deliver_reward(groups_C_D)
        agent_pos = start_pos # Reset to start
else: # Hit a wall
    deliver_punishment(groups_C_D)
    # Agent does not move if it tries to enter a wall
\end{lstlisting}

\item \textbf{Dynamic Environments and Adaptation:} To probe the adaptability of the learned world model, the environment can be made non-stationary. The boundary of the safe zone could slowly shift over the course of an experiment, or the goal location in a maze could change after a block of successful trials. This forces the agent to extinguish a previously learned policy and acquire a new one, providing a direct assay for the flexibility and update speed of its internal model (see Listing \ref{lst:avoidance_dynamic}).

\begin{lstlisting}[caption=Pseudocode for dynamic avoidance task, captionpos=b, label=lst:avoidance_dynamic]
Initialize environment with a moving boundary
agent_position = 4
safe_zone_boundary = 5 # Initially, positions 1-5 are safe
boundary_shift_interval = 100 # Timesteps before boundary moves
timestep = 0

while experiment_running:
# 1. State representation & 2. Action decoding (same as 1D)
encode_position_as_stimulus(groups_C_D, agent_position)
agent_position = update_position_from_spikes(groups_A_B, agent_position)

# 3. Feedback based on the DYNAMIC boundary
if agent_position <= safe_zone_boundary:
    deliver_reward(groups_C_D)
else:
    deliver_punishment(groups_C_D)
    
# 4. Update the environment itself
timestep += 1
if timestep % boundary_shift_interval == 0:
    # Shift the boundary occasionally, forcing the agent's world model to adapt
    if random.choice([True, False]):
        safe_zone_boundary = max(1, safe_zone_boundary - 1)
    else:
        safe_zone_boundary = min(safe_zone_boundary + 1, 8)
\end{lstlisting}

\end{itemize}

\subsection{Environment 2: Goal-seeking in a predator-prey scenario}
\begin{figure}[htbp]
\centering
\includegraphics[width=0.9\linewidth]{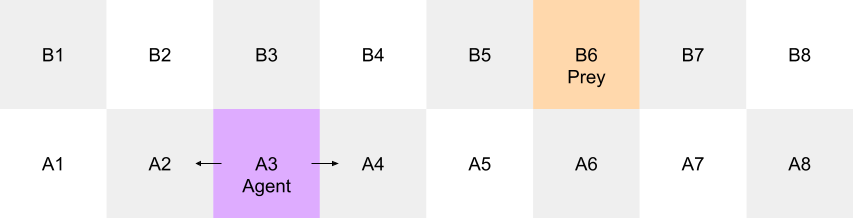} % Placeholder figure file
\caption{Diagram of the 1D Predator-Prey environment. The agent (predator) must interpret sensory information about its own location and the prey's location to decide on an action. A successful capture is rewarded, driving the agent to build a world model that supports goal-directed navigation.}
\label{fig:figure2}
\end{figure}

This task elevates the cognitive demand by requiring the agent (predator) to actively seek a dynamic goal (prey) rather than merely avoiding a static negative state. This environment simulates the fundamental evolutionary pressures of foraging, which necessitates the development of efficient world models for locating and capturing resources under uncertainty \citep{stephens1986foraging}. A diagram is shown in Figure \ref{fig:figure2}.

\subsubsection{Protocol description}
\begin{enumerate}[label=\arabic*., wide, labelindent=0pt]
\item \textbf{State Representation and Sensory Encoding:} The world state is defined by the positions of both the predator and the prey on separate 8-position grids. The agent receives this information via two distinct sensory channels: the prey's location (exteroceptive information) is encoded by stimulating a specific electrode in group C, while the agent's own location (proprioceptive information) is encoded via group D. This separation provides distinct data streams for building a model of itself in relation to the goal.

\item \textbf{Action Decoding and Execution:} The motor decoding mechanism remains identical to the avoidance task, where differential activity between recording groups A and B determines left or right movement for the predator.

\item \textbf{Feedback and State Transition:} A predictable reward is delivered when the predator's position matches the prey's position (a successful capture). Upon capture, the prey respawns at a new random location, initiating a new trial. Punishment, in the form of an unpredictable stimulus, is delivered if the agent fails to capture the prey within a time limit of $Z$ timesteps, encouraging efficient policies.
\end{enumerate}

\begin{lstlisting}[caption=Pseudocode for predator-prey task, captionpos=b, label=lst:predator]
Initialize environment
predator_pos = random.choice(A1_to_A8)
prey_pos = random.choice(B1_to_B8)
time_since_reward = 0

Main experimental loop
while experiment_running:
# Optional: Update prey position for dynamic version
# prey_pos = move_prey_procedurally(prey_pos)

# 1. State representation
encode_stimulus(group_C, position=prey_pos)
encode_stimulus(group_D, position=predator_pos)

# 2. Action decoding
spikes_A = record_spikes(group_A, window=T_ms)
spikes_B = record_spikes(group_B, window=T_ms)

if spikes_A > spikes_B:
    predator_pos = max(1, predator_pos - 1)
elif spikes_B > spikes_A:
    predator_pos = min(8, predator_pos + 1)

# 3. Feedback and state transition
if predator_pos == prey_pos:
    deliver_reward(groups_C_D)
    prey_pos = random.choice(B1_to_B8) # Respawn prey
    time_since_reward = 0
else:
    time_since_reward += 1

if time_since_reward >= Z:
    deliver_punishment(groups_C_D)
    prey_pos = random.choice(B1_to_B8) # Respawn prey
    time_since_reward = 0
\end{lstlisting}

\subsubsection{Environmental scaling and multi-agent systems}
\begin{itemize}
\item \textbf{Dimensionality and Search Strategy:} Scaling the world to 2D or 3D transforms the task from a simple directional problem into a genuine search problem. The agent must learn an efficient search strategy (e.g., patterned, memory-guided, or random) to locate the prey in a vastly larger state space. This places significant demands on the world model's capacity to integrate sensory information over extended time horizons to infer the target's location. See Listing \ref{lst:predator_2d}.

\begin{lstlisting}[caption=Pseudocode for 2D predator-prey task, captionpos=b, label=lst:predator_2d]
Initialize a 10x10 grid environment
GRID_SIZE = 10
predator_pos = (random.randint(0, 9), random.randint(0, 9))
prey_pos = (random.randint(0, 9), random.randint(0, 9))

while experiment_running:
# 1. State representation (2D)
# Encode prey position using spatial location on group C and frequency for y-axis
encode_stimulus(group_C, axis='x', value=prey_pos[0])
encode_stimulus(group_C, axis='y', value=prey_pos[1], use_freq_mod=True)
# Encode predator position similarly on group D
encode_stimulus(group_D, axis='x', value=predator_pos[0])
encode_stimulus(group_D, axis='y', value=predator_pos[1], use_freq_mod=True)

# 2. Action decoding (4-way movement)
action = decode_4way_movement(groups_A_B)
predator_pos = update_2d_position(predator_pos, action, GRID_SIZE)

# 3. Feedback for 2D capture
if predator_pos == prey_pos:
    deliver_reward(groups_C_D)
    prey_pos = (random.randint(0, 9), random.randint(0, 9))
\end{lstlisting}

\item \textbf{Introduction of an Adversary:} A second mobile entity, an adversary, can be introduced, creating a multi-objective task: approach prey while avoiding the adversary. The adversary's location would be encoded on a separate sensory channel (e.g., via a distinct stimulation frequency). Its behavior could range from a random walk to actively tracking the agent, requiring the agent's world model to account for multiple dynamic entities with different behavioral characteristics. See Listing \ref{lst:predator_adversary}.

\begin{lstlisting}[caption=Pseudocode for predator-prey with an adversary, captionpos=b, label=lst:predator_adversary]
Add a third entity, the adversary
predator_pos = 5
prey_pos = 2
adversary_pos = 8
DANGER_RADIUS = 2

while experiment_running:
# Adversary moves with a simple policy (e.g., towards predator)
adversary_pos = move_adversary(adversary_pos, predator_pos)

# 1. State representation (includes adversary location)
encode_stimulus(group_C, position=prey_pos, pattern='prey')
encode_stimulus(group_D, position=predator_pos, pattern='self')
# Use different frequency/amplitude to encode adversary on a sensory channel
encode_stimulus(group_C, position=adversary_pos, pattern='adversary')

# 2. Action decoding (same as 1D)
predator_pos = update_position_from_spikes(groups_A_B, predator_pos)

# 3. Feedback with dual objective
if abs(predator_pos - adversary_pos) < DANGER_RADIUS:
    deliver_punishment(groups_C_D, amplitude='high')
    predator_pos = random.choice(range(1,9)) # Reset on capture
elif predator_pos == prey_pos:
    deliver_reward(groups_C_D)
    prey_pos = random.choice(range(1,9))
\end{lstlisting}

\item \textbf{Multi-Organoid Systems:} The framework can be extended to create a simple competitive ecosystem. One organoid could control the prey and a second could control the predator, allowing their policies to co-evolve. This can be further extended to a linear food chain (Organoid A is prey to B, B is prey to C) or a cyclical model, in providing a novel platform to study the emergence of complex, hierarchical strategies in a multi-agent biological system. A pseudocode implementation for a two-organoid system is available in Listing \ref{lst:predator_multi}.

\begin{lstlisting}[caption=Pseudocode for multi-organoid predator-prey, captionpos=b, label=lst:predator_multi]
Initialize two separate MEA platforms
organoid_predator = MEA_Platform(id=1)
organoid_prey = MEA_Platform(id=2)

Initialize shared game state
predator_pos = 5
prey_pos = 3

while experiment_running:
# PREDATOR ORGANOID'S TURN
# 1a. Send state to predator organoid
organoid_predator.encode_state(predator_pos, prey_pos)
# 2a. Get action from predator organoid
predator_action = organoid_predator.decode_action()
predator_pos = update_position(predator_pos, predator_action)

# PREY ORGANOID'S TURN
# 1b. Send state to prey organoid
organoid_prey.encode_state(prey_pos, predator_pos)
# 2b. Get action from prey organoid
prey_action = organoid_prey.decode_action()
prey_pos = update_position(prey_pos, prey_action)

# 3. Deliver feedback to both based on outcome
if predator_pos == prey_pos:
    organoid_predator.deliver_reward()
    organoid_prey.deliver_punishment()
    # Reset positions
    predator_pos, prey_pos = 5, 3
\end{lstlisting}

\end{itemize}

\subsection{Environment 3: Dynamic interception task (Pong)}
\begin{figure}[htbp]
\centering
\includegraphics[width=0.4\linewidth]{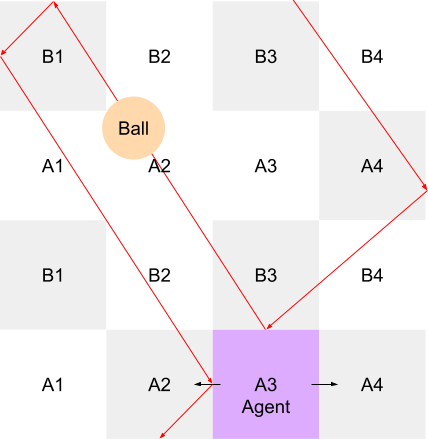} % Placeholder figure file
\caption{The Pong environment. The organoid receives complex, multi-modal sensory input about the ball's position (rate-coded distance, spatially-coded location) and must learn a predictive world model of the ball's trajectory to control its paddle and achieve a successful interception.}
\label{fig:figure3}
\end{figure}

This experiment, based on the task from \citep{kagan2022vitro}, serves as a benchmark for goal-directed behavior in a more complex, continuous-time state space. Success in Pong requires the agent to build a predictive world model of the ball's trajectory and to generate precisely timed actions to intercept it. The experimental layout is depicted in Figure \ref{fig:figure3}.

\subsubsection{Protocol description}
\begin{enumerate}[label=\arabic*., wide, labelindent=0pt]
\item \textbf{Game State Update:} The simulation first updates the ball's position based on its current velocity vector, modeling the basic physics of momentum and collision within the 2D game world.

\item \textbf{State Representation and Sensory Encoding:} The continuous 2D position of the ball is conveyed to the organoid through a rich, multi-modal encoding scheme. The ball's horizontal location (X-position) is spatially encoded, where discrete regions of the X-axis map to specific stimulation electrodes across groups C and D. Concurrently, the ball's vertical distance from the paddle (Y-position) is rate-coded, with stimulation frequency being inversely proportional to distance (e.g., high frequency when close, low frequency when far). This complex sensory input challenges the agent's world model to integrate and interpret two distinct, continuous data streams to infer the ball's state vector.

\item \textbf{Action Decoding and Execution:} The agent controls the vertical movement of its paddle. Higher aggregate spike rates from recording group A over a short time window (e.g., 10 ms) move the paddle up, while higher rates from group B move it down. This provides a continuous-like control mechanism.

\item \textbf{Feedback:} A successful interception (paddle-ball collision) results in a predictable reward, reinforcing the predictive model and the action policy. A failure to intercept (a miss) triggers an unpredictable punishment, signaling a failure in the model's prediction or the agent's control.
\end{enumerate}

\begin{lstlisting}[caption=Pseudocode for pong task, captionpos=b, label=lst:pong]
Initialize game state and MEA interface
game = PongGame()

Main game loop
while game.is_active:
# 1. Update game state
game.update_ball_position()
ball_x, ball_y = game.get_ball_position()

# 2. State representation
encode_y_pos_rate_coded(groups_C_D, ball_y)
encode_x_pos_spatial(groups_C_D, ball_x)

# 3. Action decoding
spikes_A = record_spikes(group_A, window=10_ms)
spikes_B = record_spikes(group_B, window=10_ms)

if spikes_A > spikes_B:
    game.move_paddle_up()
elif spikes_B > spikes_A:
    game.move_paddle_down()
    
# 4. Feedback
if game.check_collision():
    deliver_reward_predictable(groups_C_D)
    game.ball_bounce()
elif game.check_miss():
    deliver_punishment_unpredictable(groups_C_D)
    game.reset_ball()
\end{lstlisting}

\subsubsection{Environmental scaling and generalization}
\begin{itemize}
\item \textbf{Generalization to Other Games:} The core paradigm of Pong can be adapted to other classic arcade games to probe different cognitive functions and world modeling capabilities. For instance, \textit{Breakout} would require the agent's world model to incorporate the same ball and paddle, as well as a static array of targets, demanding precise spatial targeting to aim the ball. \textit{Space Invaders} would introduce the complexities of dodging projectiles while timing offensive actions, requiring the world model to track multiple independent agents. A pseudocode implementation for Breakout is provided in Listing \ref{lst:breakout}.

\begin{lstlisting}[caption=Pseudocode for breakout task, captionpos=b, label=lst:breakout]
Initialize Breakout game state
game = BreakoutGame() # Includes paddle, ball, and bricks

while game.is_active:
# 1. Update game state (ball movement, etc.)
game.update_ball_position()

# 2. State representation
# Same as Pong: encode ball x (spatial) and y (rate-coded)
encode_ball_state(groups_C_D, game.get_ball_position())
# Could add info about brick layout (e.g., density in ball's path)

# 3. Action decoding (paddle movement, now horizontal)
spikes_A = record_spikes(group_A, window=10_ms)
spikes_B = record_spikes(group_B, window=10_ms)
if spikes_A > spikes_B:
    game.move_paddle_left()
elif spikes_B > spikes_A:
    game.move_paddle_right()
    
# 4. Feedback
collision_type = game.check_collisions()
if collision_type == 'BRICK':
    deliver_reward_predictable(groups_C_D)
    game.remove_brick()
    game.ball_bounce()
elif collision_type == 'PADDLE':
    # Neutral or small reward for just keeping ball in play
    game.ball_bounce()
elif collision_type == 'MISS':
    deliver_punishment_unpredictable(groups_C_D)
    game.reset_ball()
\end{lstlisting}

\item \textbf{Competitive Multi-Organoid Systems:} Pong provides an ideal testbed for multi-agent competition. Two distinct organoid-MEA systems can be linked to the same game instance, each controlling one of the paddles. This setup facilitates the direct study of competitive co-adaptation, where each biological agent must build a world model that includes the policy of its opponent. The evolution of rally lengths, shot patterns, and win/loss rates would serve as powerful metrics for inter-agent learning and strategic model formation (see Listing \ref{lst:pong_multi}).
\end{itemize}

\begin{lstlisting}[caption=Pseudocode for two-organoid competitive pong, captionpos=b, label=lst:pong_multi]
Initialize two organoid interfaces and one shared game
organoid_1 = MEA_Platform(id=1)
organoid_2 = MEA_Platform(id=2)
game = PongGame(two_player=True)

while game.is_active:
# 1. Update game state
game.update_ball_position()
ball_pos = game.get_ball_position()

# 2. Send state to both organoids
# Each organoid receives the same world state information
organoid_1.encode_ball_state(ball_pos)
organoid_2.encode_ball_state(ball_pos)

# 3. Decode actions from both organoids simultaneously
action_1 = organoid_1.decode_paddle_movement()
action_2 = organoid_2.decode_paddle_movement()
game.move_paddle(player=1, action=action_1)
game.move_paddle(player=2, action=action_2)

# 4. Check for outcomes and deliver feedback independently
if game.check_paddle1_miss():
    organoid_1.deliver_punishment()
    organoid_2.deliver_reward() # Reward for opponent's miss
    game.reset_ball(server=2)
elif game.check_paddle2_miss():
    organoid_2.deliver_punishment()
    organoid_1.deliver_reward()
    game.reset_ball(server=1)
elif game.check_paddle_hit():
    game.ball_bounce()
    # Optional: deliver small reward to hitting player to encourage long rallies
\end{lstlisting}

\subsection{Generative models for world generation: LLM-driven automated curriculum design}
To overcome the limitations of manual experimental design and to systematically explore the vast parameter space of environments and training protocols, we propose an automated framework where an LLM functions as a meta-controller or AI Dungeon Master. This approach directly aligns with the theme of leveraging generative models and LLMs, applying the LLM as a tool to orchestrate the generation of the agent's training world and curriculum. This creates a high-throughput system for discovering optimal training strategies and environmental configurations that might be non-obvious to human researchers. A pseudocode implementation of this meta-learning loop is provided in Listing \ref{lst:llm}.

\subsubsection{Framework description}
The system operates in a closed loop, iteratively generating, executing, and refining experimental protocols. A central dataset logs all information from each iteration, including the LLM-generated scripts, execution outputs, agent performance metrics (e.g., capture rate, rally length), neurophysiological responses, and any errors encountered.

\begin{enumerate}[label=\arabic*., wide, labelindent=0pt]
\item \textbf{Prompt Formulation:} At the start of each iteration, a detailed prompt is programmatically constructed. This prompt provides the LLM with all necessary context, including: (1) the high-level scientific objective (e.g., "Design a protocol to increase the agent's adaptability to a moving prey"); (2) a comprehensive API specification of available commands and their valid parameter ranges, covering electrical stimulation, microfluidics, and optogenetics; (3) the current physiological state of the organoid (e.g., baseline firing rates, plasticity measurements); and (4) a summary of the historical dataset, providing examples of past protocols that succeeded or failed.

\item \textbf{Protocol Generation:} The LLM processes the prompt and generates a novel experimental protocol. This can be done in two primary modes: generating a structured JSON object with specific experimental variables to be plugged into a template script, or generating an entire Python script from scratch that defines novel experimental logic and curriculum stages. This generative capacity allows the LLM to move beyond simple parameter tuning to design entirely new tasks. Detailed examples are provided in Section \ref{sec:llm_examples}.

\item \textbf{Validation and Execution:} The generated protocol undergoes automated validation. A parser checks for syntactic correctness and ensures that all specified parameters fall within predefined safety boundaries. Invalid protocols are logged as failures, and the loop continues. Valid protocols are executed on the MEA platform, controlling the closed-loop interaction with the organoid.

\item \textbf{Data Logging and Analysis:} All data from the experimental run, the generated algorithm, the organoid's neural response, task performance metrics, and the resulting change in physiological state, are logged to the central dataset. This creates a rich, longitudinal record of the agent's learning journey.

\item \textbf{Iterative Refinement of the Generative Model:} The process repeats. Periodically, the collected dataset is used to refine the LLM's protocol generation strategy. This can be achieved through:
    \begin{itemize}
        \item \textbf{Prompt Engineering:} The base prompt is updated with successful examples from the dataset for few-shot learning, or with explicit instructions derived from common failure modes, improving the LLM's performance in subsequent iterations.
        \item \textbf{Fine-tuning:} With a sufficiently large dataset, the model can be fine-tuned on successful prompt-protocol pairs. This process trains the LLM to internalize the features of effective experimental design, creating a powerful feedback loop where the LLM becomes an increasingly sophisticated and specialized scientific discovery tool.
    \end{itemize}
\end{enumerate}

\begin{lstlisting}[caption=Pseudocode for LLM-driven experiment automation, captionpos=b, label=lst:llm]
Initialize dataset and load base prompt template
dataset = initialize_database()
prompt_template = load_base_prompt("config/llm_prompt.txt")

Main automation loop
for iteration in range(NUM_ITERATIONS):
# 1. Formulate prompt with historical data
current_organoid_state = get_organoid_baseline()
prompt = formulate_prompt(prompt_template, dataset, current_organoid_state)

# 2. LLM generates a new experimental script
generated_script = call_llm_api(prompt)

# 3. Validate the generated script for safety and syntax
is_valid, error_msg = validate_protocol_script(generated_script)

# 4. Execute or log error
if is_valid:
    # Run the experiment on the MEA platform
    results = execute_script_on_platform(generated_script)
    
    # Store the script and its outcomes in the dataset
    store_results(dataset, script=generated_script, results=results)
else:
    # Log the invalid script and the reason for failure
    store_error(dataset, script=generated_script, error=error_msg)
    
# 5. Periodically refine the LLM's instructions
if (iteration + 1) % BATCH_SIZE == 0:
    prompt_template = refine_prompt_based_on_results(dataset)
    # Optional: Trigger fine-tuning if sufficient data is collected
    # if len(dataset) > FINETUNE_THRESHOLD:
    #     launch_finetuning_job(dataset)
\end{lstlisting}

\subsubsection{LLM prompting examples}
\label{sec:llm_examples}
This section provides detailed examples of the two primary modes for using an LLM to generate experimental protocols: structured JSON output for parameter tuning and full Python script generation for implementing novel logic and curriculum design.

\paragraph{Example 1: JSON-based parameter optimization}
This approach is ideal for systematic exploration of a known parameter space. Constraining the LLM to output a JSON object ensures the protocol is syntactically correct and can be easily parsed and validated. This method is useful for optimizing feedback mechanisms by exploring different stimulation modalities and their parameters. The prompt (Listing \ref{lst:json_prompt_appendix}) asks the LLM to propose new parameters for the predator-prey task, taking into account historical data and a new hypothesis.
\begin{lstlisting}[language=, caption={Prompt for LLM to generate JSON parameters, including reward modality.}, label=lst:json_prompt_appendix]
You are an AI assistant optimizing a neuroscience experiment.
Goal: Increase the prey capture rate for a biological agent. Current rate is 25%.

Historical Data:
- Run 1: Electrical reward (20Hz, 2uA) -> 22% capture rate.
- Run 2: Electrical reward (40Hz, 2uA) -> 24% capture rate.
- Run 3: Dopamine uncaging reward (500ms duration) -> 25% capture rate, but learning is slow.

Hypothesis: A stronger, more salient electrical reward might be more effective. Let's try a tri-phasic pulse to ensure charge balance and a higher amplitude.

Propose a new set of parameters. Your output must be a valid JSON object.
Parameter constraints:
- reward_modality: one of ["electrical", "dopamine_uncaging"]
- electrical_params.shape: one of ["bi-phasic", "tri-phasic"]
- electrical_params.amplitude_uA: float, range [0.1, 20.0]
- electrical_params.pulse_duration_us: int, range [50, 500]
- dopamine_params.uncaging_duration_ms: int, range [100, 1000]
\end{lstlisting}

An example of an LLM response is shown in Listing \ref{lst:json_output_appendix}. The model correctly interprets the hypothesis, selects the electrical modality, and populates the nested parameters accordingly, including the requested tri-phasic shape and a significantly higher amplitude. This structured output can be directly loaded by the experimental control software.
\begin{lstlisting}[language=, caption={Example of a JSON output from the LLM with nested parameters.}, label=lst:json_output_appendix]
{
  "reward_modality": "electrical",
  "electrical_params": {
    "shape": "tri-phasic",
    "amplitude_uA": 8.5,
    "pulse_duration_us": 150,
    "frequency_hz": 30
  },
  "dopamine_params": null,
  "punishment_params": {
    "shape": "bi-phasic",
    "amplitude_uA": 10.0,
    "pulse_duration_us": 200
  }
}
\end{lstlisting}

\paragraph{Example 2: Full script generation for curriculum design}
This approach grants the LLM maximum flexibility to design novel experimental logic, suitable for tasks like curriculum design where the protocol itself must change based on agent performance. By providing the LLM with a detailed API specification, it can compose complex sequences of operations. The prompt in Listing \ref{lst:script_prompt_appendix} asks the LLM to write a complete Python script for a block of trials for a hypothetical platform. This includes defining the logic for state encoding, action decoding, feedback delivery, and a terminal condition with a suggestion for the next curriculum step. This example uses FinalSpark's Neuroplatform \citep{jordan2024open}.
\begin{lstlisting}[language=, caption={Prompt for LLM to generate a full Python script with a detailed API.}, label=lst:script_prompt_appendix]
You are an AI assistant designing a Python script for the Neuroplatform.
Task: Write a script for one block of the predator-prey experiment.
Goal: Teach the agent to capture a stationary prey.

Use the provided NeuroplatformAPI. Key functions:
- api.encode_spatial(electrode_num): Stimulate one electrode for location.
- api.get_spike_counts(electrodes_A, electrodes_B): Returns (count_A, count_B).
- api.fire_pulse_train(electrodes, shape, amplitude_uA, pulse_us, freq_hz, duration_ms): Delivers electrical feedback.
- api.uncage_dopamine(duration_ms): Delivers chemical reward.

Write a Python script for a 10-trial block.
- Prey is stationary at position 6 (encoded on electrode 14).
- Predator starts at position 2 (encoded on electrode 2).
- Reward: A 500ms pulse train at 40Hz on all sensory electrodes.
- Punishment: A high-amplitude (15uA) single bi-phasic pulse.
At the end, print the capture rate and suggest a curriculum step. If capture rate > 70%, suggest making the prey move. Otherwise, suggest repeating.
\end{lstlisting}

The resulting script (Listing \ref{lst:script_output_appendix}) demonstrates the LLM's ability to correctly use the specified API. It structures the trial loop, implements the state encoding and action decoding logic, calls the `fire\_pulse\_train` function with correct parameters for reward and punishment, and prints a context-aware suggestion, effectively automating a step in curriculum design.
\begin{lstlisting}[caption={Example of a Python script generated by the LLM using a detailed API.}, label=lst:script_output_appendix]
from neuroplatform_api import NeuroplatformAPI
import time
import random

def run_predator_prey_block(api):
    num_trials = 10
    captures = 0
    
    # Define sensory and motor electrodes
    SENSORY_ELECTRODES = [10, 11, 12, 13, 14, 15, 16, 17]
    MOTOR_A_ELECTRODES = [0, 1] 
    MOTOR_B_ELECTRODES = [8, 9]

    for trial in range(num_trials):
        predator_pos = 2
        prey_pos = 6
        
        for timestep in range(15): # Max 15 steps per trial
            api.encode_spatial(electrode_num=prey_pos + 8) # Prey on group D
            api.encode_spatial(electrode_num=predator_pos) # Predator on group C
            time.sleep(0.1)

            spikes_A, spikes_B = api.get_spike_counts(MOTOR_A_ELECTRODES, MOTOR_B_ELECTRODES)
            
            if spikes_A > spikes_B:
                predator_pos = max(1, predator_pos - 1)
            elif spikes_B > spikes_A:
                predator_pos = min(8, predator_pos + 1)
            
            if predator_pos == prey_pos:
                # Deliver complex electrical reward
                api.fire_pulse_train(
                    electrodes=SENSORY_ELECTRODES, shape='bi-phasic', 
                    amplitude_uA=5.0, pulse_us=200, 
                    freq_hz=40, duration_ms=500
                )
                captures += 1
                break
        
        if predator_pos != prey_pos: # Punishment if not captured in time
            api.fire_pulse_train(
                electrodes=random.sample(SENSORY_ELECTRODES, 4), shape='bi-phasic',
                amplitude_uA=15.0, pulse_us=200,
                freq_hz=0, duration_ms=1 # Single pulse
            )

    capture_rate = captures / num_trials
    print(f"Block finished. Capture rate: {capture_rate:.2f}")
    if capture_rate > 0.7:
        print("Suggestion for next block: Success. Make prey move predictably.")
    else:
        print("Suggestion for next block: Low performance. Repeat block with increased reward saliency.")

if __name__ == '__main__':
    platform_api = NeuroplatformAPI()
    run_predator_prey_block(platform_api)
\end{lstlisting}

\section{Discussion}
The framework presented in this paper resides at the confluence of embodied AI, model-based reinforcement learning, and computational neuroscience. By treating a neural organoid as an embodied agent that must learn an internal world model, we open up new avenues for both fundamental scientific inquiry and AI research.

\subsection{Broader implications and future work}
This work represents a paradigm shift from viewing organoids as passive models of neural circuitry to active agents engaged in goal-directed decision-making. This reframes core questions from "What does this neural activity represent?" to "What predictive model is this agent using to achieve its goals?" This perspective aligns seamlessly with reinforcement learning and predictive coding theories like the Free Energy Principle, suggesting a shared computational language for biological and artificial intelligence. The ability to directly measure the physical implementation of the world model via synaptic plasticity provides a ground truth for learning that is unavailable in purely \textit{in silico} systems.

The logical progression of this research involves scaling the complexity of both the agent's embodiment and its environment:
\begin{itemize}
\item \textbf{Environment Scaling:} Moving beyond the simple 2D tasks presented here to more complex, physics-based simulations, partially observable environments, or multi-agent scenarios where organoids can cooperate or compete with each other or with traditional AI agents.
\item \textbf{Interface Scaling:} Increasing the communication bandwidth between the agent and its world through high-density MEAs or integrating optogenetics for cell-type-specific stimulation and recording. A richer I/O channel would enable the agent to form a more detailed world model and execute a more expressive range of actions.
\item \textbf{Sim-to-Bio-to-Real Embodiment:} A crucial future direction is to bridge the \textit{in vitro}-to-physical gap. By connecting these biological agents to simple robotic actuators (e.g., a wheeled robot) and sensors (e.g., a camera), we could create truly embodied agents that must learn a world model of physical reality. This provides a powerful new platform for studying grounded cognition and tackling sim-to-real challenges from a novel biological perspective, directly addressing aligning simulation with real-world physics.
\end{itemize}

\subsection{Limitations and challenges}
Despite its significant promise, this approach has its challenges. The biological substrate introduces inherent variability, limited lifespans, and developmental stochasticity, making reproducibility a central concern that must be addressed with robust experimental design and statistical analysis. This variability arises from multiple sources, including genetic and epigenetic differences between iPSC lines \citep{Mancinelli2025emergence} and the stochastic nature of self-organization, particularly in unguided protocols \citep{Lancaster2014Generation}. The limited lifespans and long-term health of large organoids are also a practical constraint, as the tissue's non-vascularized core is susceptible to hypoxia and necrosis, which can introduce experimental artifacts \citep{Zhang2022}. The organoid itself remains a profound black box; while we can measure network-level plasticity, deciphering the specific computational algorithms it implements to learn and represent its world model is a formidable open problem. Furthermore, the information bandwidth of current MEA technology, while improving, is still orders of magnitude lower than that of biological nervous systems, potentially constraining the complexity of the world models that can be learned.

\section{Conclusion}
This paper has introduced a comprehensive framework for the design, scaling, and evaluation of virtual environments for a new class of embodied agent: the human neural organoid. By structuring tasks as a coherent curriculum of increasing complexity, from conditional avoidance to goal-seeking and dynamic interception, we can systematically probe and induce the formation of internal world models. Our proposed methods for sensory encoding, motor decoding, and feedback are deeply grounded in the principles of computational neuroscience and model-based reinforcement learning.

Our key contributions are threefold, each directly relevant to the study of embodied world models. First, we provide a concrete set of scalable environment designs tailored to induce and examine learning in biological neural networks. Second, we introduce a novel evaluation paradigm that links an agent's behavioral performance directly to the physical measurement of synaptic plasticity, offering an unprecedented view into the material basis of a learned world model. Third, we propose a forward-looking, LLM-driven generative methodology for automating and scaling the design of training curricula, which holds broad implications for both Organoid Intelligence and the future of AI-driven scientific discovery. This work represents a critical step toward creating more sophisticated and adaptive biological agents, offering a unique and powerful platform to explore the fundamental connections between environment, embodiment, and intelligence.

\begin{ack}
We thank FinalSpark for suggesting the investigation of LLM based automated research.
\end{ack}

\bibliographystyle{plainnat}
\bibliography{main}

\end{document}